\documentclass[letterpaper]{article} 
\usepackage[draft]{aaai25}  
\usepackage{times}  
\usepackage{helvet}  
\usepackage{courier}  
\usepackage[hyphens]{url}  
\usepackage{graphicx} 
\usepackage{natbib}  
\usepackage{caption} 
\usepackage{algorithm}
\usepackage{algorithmic}

\usepackage{newfloat}
\usepackage{listings}
\DeclareCaptionStyle{ruled}{labelfont=normalfont,labelsep=colon,strut=off} 
\floatstyle{ruled}
\newfloat{listing}{tb}{lst}{}
\floatname{listing}{Listing}
\usepackage{amsmath}
\usepackage{changepage}
\usepackage{booktabs}  
\usepackage{multirow}  
\usepackage{subfig}
\usepackage{subcaption}
\usepackage{soul}

\usepackage[utf8]{inputenc}
\usepackage[breakable]{tcolorbox}
\usepackage{enumitem}
\usepackage{lipsum}

\usepackage{xcolor} 
\definecolor{SDEblue}{RGB}{28, 58, 88}

\title{Towards Neural Network based Cognitive Models of \\Dynamic Decision-Making by Humans}
\author{
    Changyu Chen\equalcontrib$^{1}$,
    Shashank Reddy Chirra\equalcontrib$^{1}$,
    Maria José Ferreira\equalcontrib$^{2}$, \\
    Cleotilde Gonzalez$^{2}$, 
    \textbf{Arunesh Sinha$^{3}$,
    Pradeep Varakantham}$^{1}$
}
\affiliations{
    $^{1}$Singapore Management University\\
    $^{2}$Carnegie Mellon University,
    $^{3}$Rutgers University\\
    \texttt{\{cychen.2020,shashankc,pradeepv\}@smu.edu.sg;} \\
    \texttt{mariajor@andrew.cmu.edu; coty@cmu.edu; arunesh.sinha@rutgers.edu}
}

\usepackage{bibentry}

\begin{document}

\maketitle

\begin{abstract}
Modeling human cognitive processes in dynamic decision-making tasks has been an endeavor in AI for a long time because such models can help make AI systems more intuitive, personalized, mitigate any human biases, and enhance training in simulation. Some initial work has attempted to utilize neural networks (and large language models) but often assumes one common model for all humans and aims to emulate human behavior in aggregate.  However, the behavior of each human is distinct, heterogeneous, and relies on specific past experiences in certain tasks. For instance, consider two individuals responding to a phishing email: one who has previously encountered and identified similar threats may recognize it quickly, while another without such experience might fall for the scam. 
In this work, we build on Instance Based Learning (IBL) that posits that human decisions are based on similar situations encountered in the past. However, IBL relies on \emph{simple fixed form functions} to capture the mapping from past situations to current decisions. To that end, we propose two new attention-based neural network models to have open form non-linear functions to model distinct and heterogeneous human decision-making in dynamic settings. We experiment with two distinct datasets gathered from human subject experiment data, one focusing on detection of phishing email by humans and another where humans act as attackers in a cybersecurity setting and decide on an attack option. We conducted extensive experiments with our two neural network models, IBL, and GPT3.5, and demonstrate that the neural network models outperform IBL significantly in representing human decision-making, while providing similar 
interpretability of human decisions as IBL. Overall, our work yields promising results for further use of neural networks in cognitive modeling of human decision making. 
Our code is available at https://github.com/shshnkreddy/NCM-HDM. 
\end{abstract}

\section{Introduction}

Since the early days of AI, cognitive approaches to AI have aimed to emulate human cognitive processes with the goal of designing computer agents that exhibit human-like decision making~\citep{riedl2012interactive, gonzalez2023building} and (possibly superior) intelligence~\citep{newell1994unified}. While recent advances in AI have relied on neural networks, the goal of making realistic human agents is still very relevant. Work in this space has built human-like agents, particularly in game play~\citep{laird2001knows,miyashita2017developing,siu2021evaluation,asfour2023harnessing}. Prior approaches aim to replicate human decision making in the aggregate (e.g., have one model for all humans)  or within broad categories of human personalities (e.g., impulsive, cautious) or simple personalized models. {\color{black}However, human decision-making is complex and heterogeneous, which does not neatly fit into categories of personalities.} The need for \emph{high fidelity personalized model} of observed humans' decision-making arises in many domains such as human detection of phishing email, understanding human attackers in cyber attacks, knowledge consumption in education settings, and many others. In the application of phishing, attacks rely on heterogeneity of decision-making, including human fallacies, to be successful. Developing realistic dynamic decision-making models for a particular human from \emph{observed past decisions} can lead to the development of personalized interventions that make humans better at detecting phishing emails. In this work, we \emph{aim} to build high fidelity, fine-grained and data driven models of human behavior in \emph{specific complex tasks}. 

We base our work on a popular cognitive model of human behavior, namely Instance-Based Learning (IBL). 
IBL is a decision-making algorithm guided by the
recognition and retrieval of past experiences (instances)
by their similarity to the current decision situation \cite{gonzalez2003instance}.
An \emph{instance} is a memory unit that consists of three elements: a situation (set
of attributes that give context to the decision, or state s); a
decision (the action taken); and a feedback (utility or
experienced outcome of the action). The IBL model has been widely used in various contexts to model human behavior~\citep{gonzalez2003instance,cranford2021towards}; however, IBL uses simple fixed-form formulas to represent the mapping from past memory instances to current decision. Other works in human behavior modeling produce a monolithic model~\citep{cheung2021attacker,binzturning} or model generic behavior on many tasks~\citep{mao2023alympics,park2023generative}. In contrast, we aim to mimic diverse and complex behavior for a \emph{specific task}.

We build two different attention-based neural network models. The core cognitive idea borrowed from IBL is to personalize the model to each human by conditioning it on the memory of the human and our \emph{main novelty} is in using the rich class of neural network based functions to map the past memory units, i.e., tuples of (situation, action, feedback), and the current situation to the next decision. We posit that the neural network's ability to represent mappings with greater richness, compared to the simple fixed-form functional representation in IBL, allows our model to more accurately capture complex human behavior, resulting in higher fidelity. Our first model, Token-Level Personalized Memory Integrated Model (TL-PMIM), is based on fine-tuning a Large Language Model (LLM), namely Llama 7B, with the given limited dataset, that provides a highly expressive model but at the cost of interpretability. Our second hybrid model, Instance-Level Personalized Memory Integrated Model (IL-PMIM), balances expressivity and interpretability by building a simple attention layered model, where the attention weights provide direct interpretability of decision making. We also consider GPT3.5 as another alternative.

We conducted thorough experiments using data gathered in prior work from two separate human subject experiments: one experiment involved participants detecting phishing emails and another experiment involved participants acting as attackers in a cyber-security deception game. We analyze the results from IBL, TL-PMIM, IL-PMIM, and GPT3.5. We find that TL-PMIM best mimics individual human behavior, outperforming other approaches. 
We also show that the prior knowledge inbuilt in GPT may not always be beneficial in modeling a single human's behavior. Overall, we believe that our work at the intersection of behavioral science and AI will open doors for further research in this area.


\section{Related Work}
\textbf{Human Behavioral Modeling}: There is a vast amount of literature on modeling human behavior in different areas of research~\citep{fuchs2023modeling,kahneman2002foundations}, all of which is not possible to describe in short. IBL differs from other prominent cognitive models such as Soar~\cite{laird2022analysis} (cognitive architecture) and ACT-R (uses a production rule system), in its reliance on specific instances rather than abstract rules or distributed representations. More related, IBL and the Generalized Context Model~\cite{nosofsky2011generalized} share similarities in the use of similarity judgments by comparing new stimuli with instances in memory and considering context sensitivity. However, IBL models can be applied to general decision-making and problem-solving across various domains, including categorization and decisions from experience, while GCM models are specifically designed to explain categorization processes. Reinforcement learning (RL) also has cognitive basis, however, not without criticism~\cite{eckstein2021reinforcement}. But, the tasks considered in this work do not have the flavor of being concerned about future rewards, which is why building on IBL (that has extensive usages and is very well established on these tasks) is promising. \citet{binzturning} finetune an LLM on simple psychological experiments, unlike the real-world complex tasks considered in this work. They also lack a memory module and hence do not allow for ready personalization, instead they build a joint model and simulate a variety of behaviors by adding random effects. Similarly, \citet{xiong2023distilling} propose an RNN based model but mention a memory module as future work.


\textbf{Human Behavioral Modeling in Phishing Email Domain}:  Researchers have thoroughly investigated human behavior when tasked to identify a new (phishing or ham) email. Prior work has studied the factors that make a phishing email hard to be detected~\citep{singh2020makes}. Moreover, prior work has applied instance-based learning (IBL) in modeling human's response to the emails from observed data~\citep{cranford2019modeling,xu2022modeling}. A recent work~\citep{asfour2023harnessing} has attempted to simulate human behavior response to phishing emails using LLMs by asking the LLM to simulate broad personality traits; this work is not data driven and does not provide a personalized modeling of a particular human (based on observed past decisions) and hence such work cannot be used for developing personalized interventions for training humans to detect emails. Prior work on interventions has also mostly relied on domain expert advice~\citep{sheng2009improving,kumaraguru2010teaching} rather than building a personalized model of human sub-optimality. 

Somewhat further away, there are works that focus on generating phishing emails by LLMs~\citep{hazell2023large,karanjai2022targeted,sharma2023well} and on detecting phishing emails by LLMs~\citep{heiding2023devising}.
Some works have built autonomous agents using LLMs with memory components~\citep{mao2023alympics, park2023generative}. These works build more generic agents and study emergent behavior, unlike our goal of heterogeneous personalized models built from observed past decisions in a specific task. 

\textbf{Human Behavioral Modeling of Attackers in Cyber-security}: This is a relatively less explored research topic. Some research in this area has originated from behavioral game theory that explores beyond the rational model of human behavior~\citep{cheung2021attacker,thakoor2020exploiting} originating in Stackelberg security games~\citep{sinha2018stackelberg}. Recently, there have been applications of the IBL model in more realistic modeling of attacker behavior~\cite{cranford2021towards}. More broadly, work in physical security has explored different aspects of human behavior in games~\citep{kar2015game}.

\begin{figure*}
    \centering
    \includegraphics[width=0.91\textwidth]{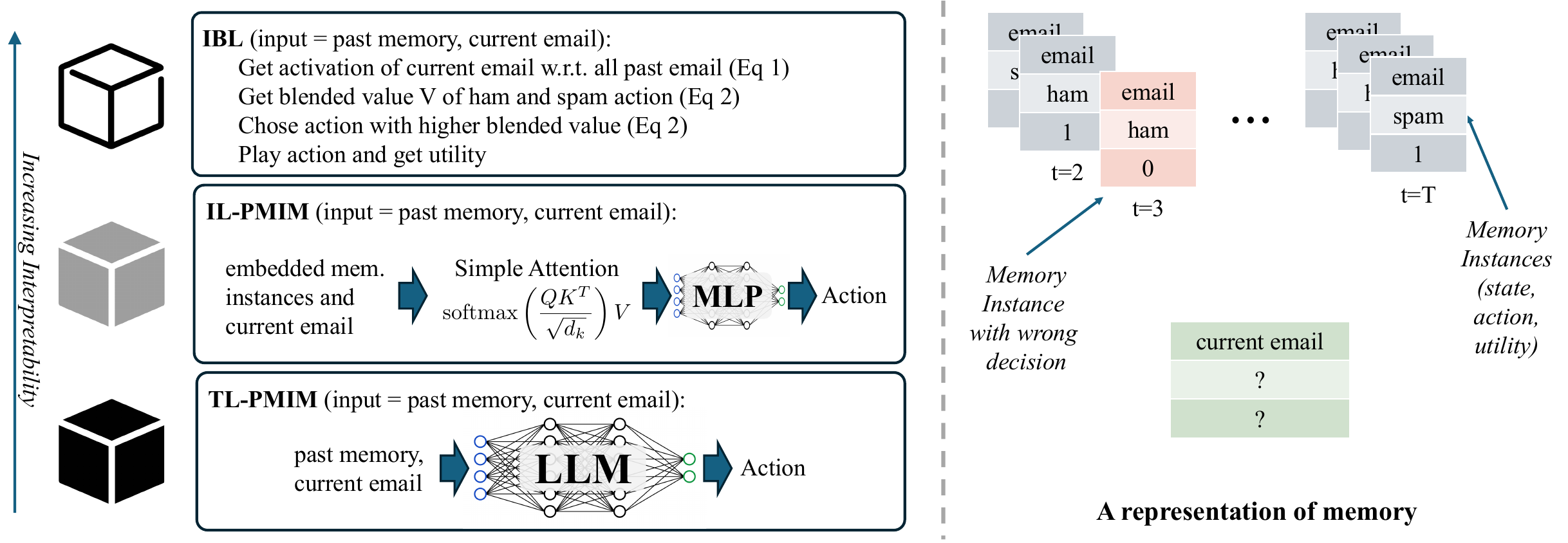}
    \caption{\color{black}\textbf{Overview of Models for the Phishing Task.} (Left) The IBL model and our newly proposed models all process inputs that consist of past memory and the current email. The interpretability of these models increases progressively from TL-PMIM to IL-PMIM and is highest in IBL. The IBL model derives interpretability from an explicit fixed-form formula, whereas IL-PMIM relies on attention weights to determine the contribution of each email. TL-PMIM, which processes each token in the email in a highly non-linear manner, offers the least interpretability. (Right) The memory maintains a set of email instances, each represented by a tuple consisting of state, action, and utility.}
    \label{fig:main}
\end{figure*}

\section{Behavioral Models with Memory Mechanism}

We describe the different approaches that we consider in this work. We start with the known IBL model in order to introduce terminology and concepts that help describe the new proposed models. 

\subsection{IBL Model} 
In Instance Base Learning~\cite{gonzalez2003instance}, a human action is modeled as the activation of memory instances tied to past experiences associated with each action or decision. A memory instance is a tuple $(s,a,u)$ denoting a situation (state) $s$ in which action/decision $a$ was taken resulting in outcome/utility of a scalar $u$. A state $s_i$ is represented by $K$ attributes, with $s_{i,k}$ denoting the $k$-th attribute. An action $a \in A$ belongs to the given set of valid actions $A$. The memory is composed of a sequence of such tuples. Memory activation of the memory instance $i$ is modulated by at least two processes (i.e., free parameters in the model): memory decay and noise associated with the retrieval of these memories. At time step $t$, let there be $n < t$ unique memory instances in memory and suppose a new state $s_t$ is encountered at the current time $t$, then the activation score $A_i(t) $ of memory instance $i$ $(s_i, a_i, u_i)$, is given by: 
%

{
\small
\begin{align}
\ln \big(  \sum_{t' \in \mathcal{T}_i(t)} (t - t')^{-d}\big) +  \mu \sum_{j =1}^K \omega_j Sim(s_{i,j},s_{t,j}) + \sigma \xi 
\label{eq:activation}
\end{align}
}%

\noindent where $d$ is a decay parameter, $\mathcal{T}_i(t)$ is the set of time steps of occurrence of instance $i$ in memory before time $t$, $Sim$ is a domain specific similarity function with range in $[0,1]$, $\mu, \omega_j > 0$ are weights (typically $1$), and noise $\sigma \xi$ introduces stochasticity in retrieval and it is a $\sigma$ scaled random variable $\xi$ with zero mean logistic distribution.
The activation score of each instance in memory determines its retrieval probability. Consider an action $a$ and the set of all instances in memory with action $a$ as $M_a$.
The \emph{retrieval probability} is calculated for each instance $i \in M_a$ using the Boltzman softmax function $P_{i,t} = \frac{e^{A_{i}(t)/\tau}}{\sum_{j \in M_a} e^{A_{j}(t)/\tau}}$ and then the model chooses the action $a^*$ with the highest \textit{blended} value \textit{V}:
\begin{equation}
V_{t}(a) = \sum_{i=1}^{|M_a|}P_{i,t}u_i \; , \; a^* = \arg\!\max_{a \in A} V_{t}(a)
\label{eq:prob_retrieval} 
\end{equation}
\noindent where $\tau$ is a temperature parameter with a default value of $\tau=\sigma\sqrt{2}$. We note that the \emph{blending} operation here is exactly like 
the soft attention weighted retrieval in modern deep learning but the blending concepts in cognitive science (IBL and ACT-R) predate these modern notions. Blending concepts are themselves inspired by similar ideas present in early AI literature~\citep{gonzalez2003instance}. Another concept used in IBL is called \emph{tracing} ~\citep{Cranford2020,Lebiere2023} inspired from feedback in intelligent tutoring systems~\citep{Anderson95}. When tracing, the model fills the memory using the true ground memory instances due to the human participant action and outcome for each new prediction, instead of the model using its own prediction and outcome. 
This aspect is analogous to teacher forcing~\cite{williams1989learning} in recurrent neural networks in AI literature.
The connections of IBL to AI concepts inspired us to explore modern neural network techniques, as described next.


\subsection{Token-Level Model}
Pretrained on extensive datasets, LLMs have demonstrated significant potential in tasks such as text generation and semantic analysis. Previous work has found that fine-tuning a pre-trained LLM enhances its ability to address new problems effectively. Following the IBL model, we incorporate personalized memory into the LLM, where this memory comprises a sequence of memory instances. This approach is called Token-Level Personalized Memory Integrated Model (TL-PMIM). This approach is implemented by integrating the memory into the context. Mathematically, we concatenate the instance sequences in the memory with the current situation as the input $s_t$ and construct the input-output pair $(x_t, a_t)$ where $x_t$ is: 
\begin{equation*}
    x_t = s_1 || a_1 || u_1 \ldots s_{t-1}|| a_{t-1} || u_{t-1} || s_t
\end{equation*}
We then fine-tune an LLM with the data $\mathcal{D}=\{(x_t, a_t)\}^N_{i=1}$. A concrete input example is provided in the appendix. 

\subsection{Instance-Level Model}
While TL-PMIM offers substantial model expressivity, it is associated with higher computational costs and reduced interpretability. To address these issues, we propose the Instance-Level Personalized Memory Integrated Model (IL-PMIM), which balances expressivity and interpretability through a simple attention-layer-based architecture. The simple attention layer provides a direct interpretation of importance in terms of the learned attention weights. 
Similar to TL-PMIM and IBL, the memory is a sequence of memory instances. However, IL-PMIM introduces an additional processing step. Specifically, each memory instance $(s,a,u)$ is encoded into two latent vectors by a text encoder $E$. One latent vector represents the state and utility, denoted by $e^{s,u}$, and the other represents the action, denoted by $e^a$. The attention layer then processes this sequence of encodings along with the encoding of the current state to produce the action for the current step. We note that the choice of producing two encoded vectors per memory instance is a trade-off: using a single latent vector for a memory instance led to the attention ignoring the action probably because likely the action information was lost in the encoding. On the other hand, using three separate encoded vectors for state, action, and utility leads to a long-memory sequence which is not processed well by the simple attention layer.  Mathematically, IL-PMIM can be expressed as follows: 

{\small
\begin{align}
    & e^{s,u}_i = E(s_i || u_i),\;
    e^a_i = E(a_i) \text{ for } i<t-1, \;
    e^s_t =E(s_t) , \nonumber\\
    & x_t = (e_1^{s,u}, e_1^a, \ldots, e_{t-1}^{s,u}, e_{t-1}^a, e^s_t), \nonumber \\
    & \{A_j\}_{j=1}^{2t-1}, \{V_j\}_{j=1}^{2t-1} = Att_w(x_t), \; \hat{a}_{t} = f_\theta(\sum_{j=1}^{2t-1} A_j V_j) \;  \label{eq:attention}
\end{align}
}

\noindent where $f_\theta$ is a multilayer perception producing the predicted action $\hat{a}_{t}$ and $Att_w$ is a simple attention layer with $w$ the weights of the query, key, and value transformation from each component of $x_t$. Note that the attention weights $A_j$ for $j\!=\!1,3,\ldots, 2t\!-\!1$ correspond to the retrieval probabilities in IBL, thereby providing interpretability similar to IBL.

{\color{black}Figure~\ref{fig:main} provides an overview of these three models, illustrated using the phishing task as an example.}






\subsection{GPT3.5 Approach} \label{sec:gptdesc}
We also experiment with the zero-shot performance of LLMs as cognitive models. Similarly to the TL-PMIM model, the GPT3.5 model is incorporated with a personalized memory module. Due to cost reasons, we cannot pass the entire memory in the prompt, hence we process the memory to retrieve the top $k$ instances. Let $s_t$ be the current email in consideration, then each instance in the memory $(s_i, a_i)$ for $i \in [1, t-1]$ is given a score $Sim(s_i, s_t)$, where $Sim$ is a domain specific similarity function. The top $k$ instances ${s_i, a_i}$  are then retrieved to create the input $x_t$ as, 
\begin{equation*}
    x_t = s_1 || a_1 || u_1 \ldots s_{k} || a_{k} || u_{k} || s_t
\end{equation*}
The LLM is then prompted to provide action $a_t$. Note that the full prompt also contains additional information such as task description and an in-context example to help the LLM. It is also asked to provide a justification for its decision to ensure that the LLM uses the memory to make the decision vs its own prior knowledge. An example of the prompt and justification given by the LLM is provided in the appendix.

\section{Experiments}\label{sec:exp}
The experiments are designed to evaluate if our approach can model human responses with high fidelity, i.e., decision pattern of our model is similar to that seen in the data. For TL-PMIM, we have fine-tuned Llama 2 7B base using LongLora~\citep{longlora}. For the encoding $E$ in IL-PMIM we use a sentence transformer, namely all-mpnet-base-v1~\citep{all-mpnet-base-v1}.

We consider two baselines from the literature. One is the IBL model~\citep{cranford2019modeling,cranford_towards_2021}.  Another is the direct use of the GPT3.5 model. 
 All neural network experiments were run on NVIDIA Quadro RTX 5000 GPUs except for TL-PMIM, which was run on NVIDIA H100 GPUs. The hyperparameters for TL-PMIM are constant learning rate 2e-5 with a warmup of 20 steps, batch size 32, 1 training epoch with AdamW optimizer. The hyperparameters for IL-PMIM are a constant learning rate $5e$-$6$ with a batch size of $64$ and a dropout of $0.1$. The key, query and value embedding dimensions are $256$. The model is trained for $1000$ epochs with an Adam optimizer.

\subsection{Domain and Data Description}

In this work, we explore two dynamic decision-making domains. The two domains chosen require somewhat varying skills: one requires visual and memory skills to detect phishing email, the other requires quantitative risk assessment skills to decide the best course of action.  Data for both domains were obtained from human subject studies in prior works. We describe these human subject studies in brief and the data gathered from these studies below.

\textbf{Phishing Dataset}: We use the Phishing dataset from~\cite{singh2019training}. The data is collected from the human participants in the human subject experiment. The experiment had three phases: pre-training, training, and post-training. Each participant received 10 emails in the pre- and post-training phases, respectively, while they received 40 emails in the training phase. The 10 post-training emails (2 phishing, 8 ham) were deliberately chosen from a different distribution to test the response to the distribution. For every email, they are asked to answer if the email is phishing or ham. These emails were selected from a pool of 239 emails, including legitimate (ham) and phishing emails.
The dataset had data from 297 participants. Three groups (same size) of participants receive three different percentages of training emails that were phishing emails: 25\%, 50\%, and 75\%. In addition to this data, we also have auxiliary data which provide six binary labels for every email used in the study. These labels are from domain experts and characterize why the email is phishing or not. The label names are self-explanatory: (1) sender mismatch, (2) request credentials, (3) subject suspicious, (4) asks to do something urgently, (5) has some offer(s), and (6) mismatch in link text and location. We do not use these auxiliary data in training, but we use them for evaluation of our model.

\begin{figure}[t]
    \centering
    \includegraphics[width=0.79\linewidth]{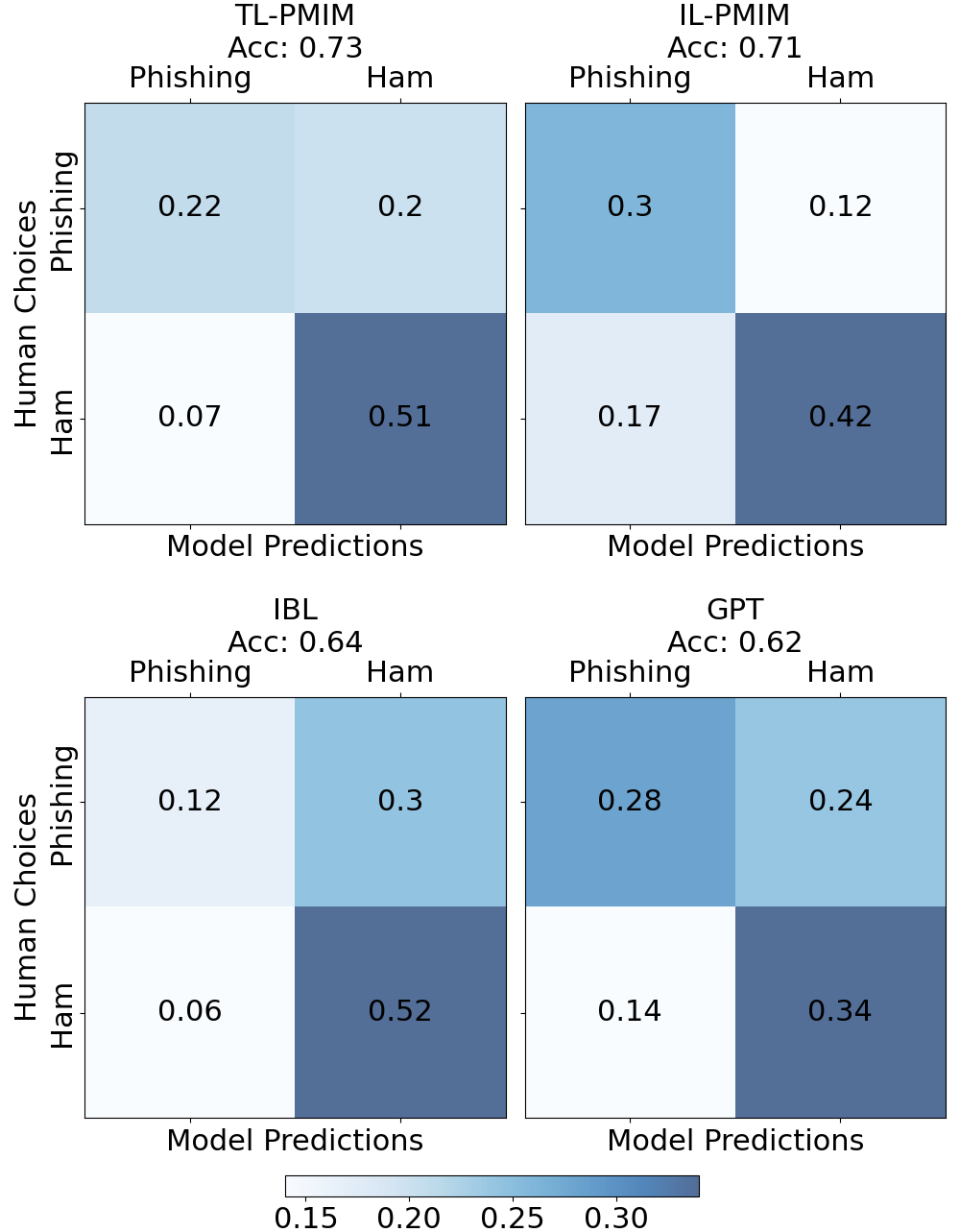}
    \caption{Phishing confusion matrix with 50:10 split}
    \label{fig:phish5010}
\end{figure}

The email information is the state for our memory instance; this state data has the sender, subject and body of email represented in words. The action is binary: yes or no for phishing. The utility is 0 or 1 based on the correct detection of phishing or ham. To investigate the effects of data size on training our models, we split the dataset in two ways. In the first way, the first 10 emails are used for training while the remaining 50 emails are used for testing (10:50 split). In the second way, the first 50 emails are used for training while the last 10 emails are used for testing (50:10 split).

As stated earlier, we need similarity scores between emails. In IBL, we followed prior work~\citep{xu2022modeling} where cosine similarity was computed on BERT embeddings. For GPT, we used an embedding tool from~\cite{ni2021large} to embed emails and then used cosine similarity.

\textbf{IAG dataset}:  The Insider Attack Game (IAG) is designed as an online game that simulates real-world scenarios in the lab such as in~\cite{cooney_learning_2020}. It involves two stages: first, defenders protect nodes (computers) by a randomized allocation of limited resources and the attacker chooses a node to attack. In the second stage, the defender chooses probabilistically to either warn the attacker that the node is defended (which could be a lie) or not provide such warning, and then the attackers choose to continue the attack or not. The defender's defense and deceptive strategies are decided based on prior work in Stackelberg games~\citep{xu2015exploring}. The game has six nodes (computers), with the resource constrained defender able to protect only two.

\begin{figure}[t]
    \centering
    \includegraphics[width=0.79\linewidth]{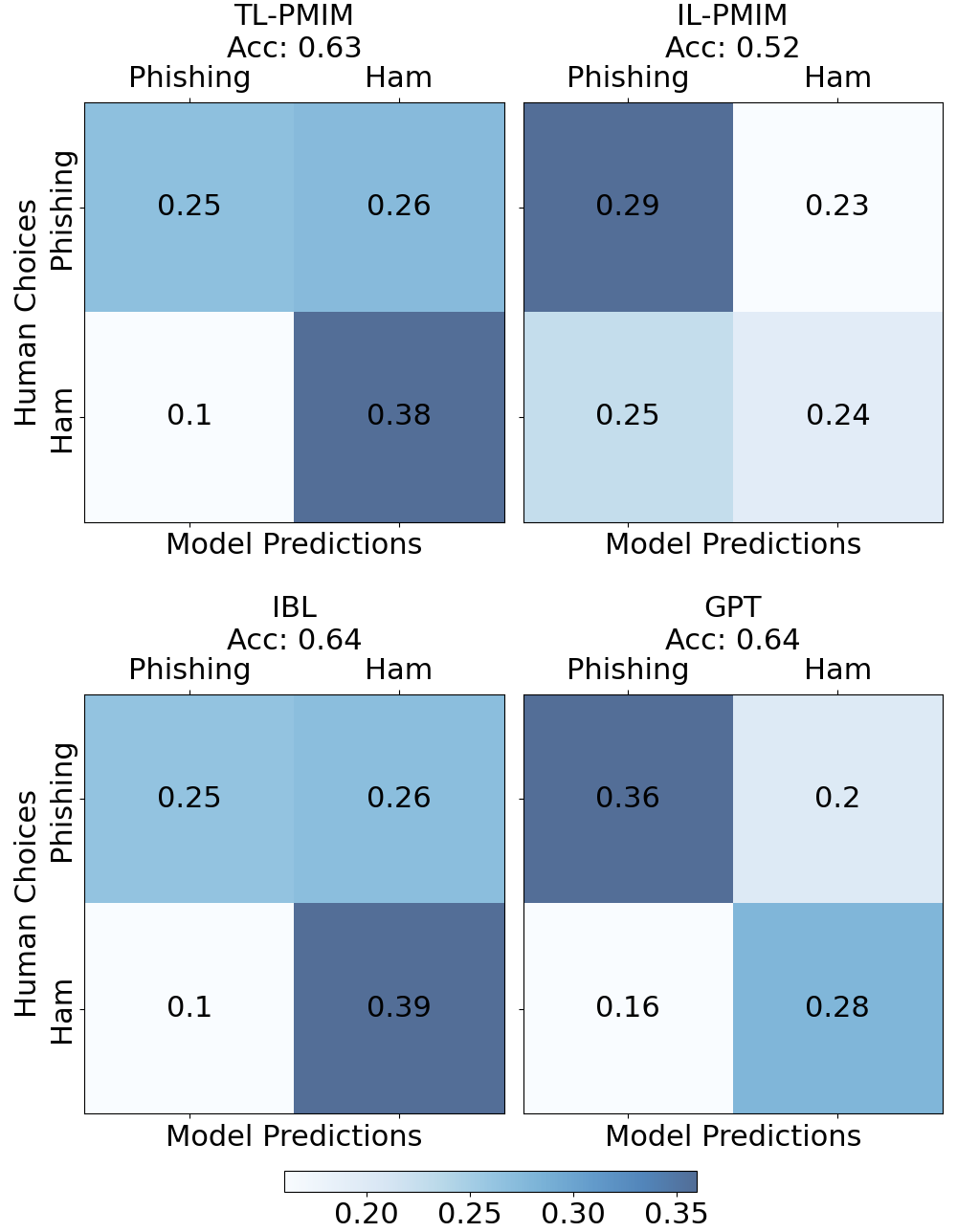}
    \caption{Phishing confusion matrix with 10:50 split}
    \label{fig:phish1050}
\end{figure}

In the human subject study, participants acted as insider attackers, aiming to gain points by attacking nodes. Attackers earn points if the node they attack is undefended but lose points if caught. The exact number of points depends on the node, ranging from -10 to +10. If an attacker withdraws the attack in the second stage, they don't earn any points. Here, we aim to model the second stage decision of the human participant, i.e., whether to continue the attack or not. The decision to continue the attack is optimal if the ground truth status of the node is uncovered (undefended) and not optimal if the node is covered.
We obtain the dataset from~\cite{aggarwal_deceptive_2023}. The dataset has information on which node was chosen (target number, target location), the probability of the node being defended, the reward (or penalty) points if the node was undefended (or defended), the actual defended or not status of the node, and the outcome expressed in the number of points obtained by the participant based on his decision to continue attack or not. In the dataset, participants played the game 100 times each for different settings (different node points, defender strategy, etc.) for a total of 107,900 game instances played. Given the larger number of data points (than the phishing dataset), we perform a 50\% split for training and testing for the IAG dataset.

As stated in earlier sections, we need similarity scores between the state features. In IBL, we follow the approach in prior work~\cite{cranford_learning_2018,nguyen_speedyibl_2023} where similarity is calculated for each feature and then summed over the features. For both reward and penalty features the similarity is $1 - \frac{\left| x - y \right|}{10}$ and for the probability of defended feature the similarity is $1 - \left| x - y \right|$. For other features (target number, target location, and defended) an exact
match is $1$ or else $0$.
As described in GPT3.5 approach, we use the same similarity score to trim the prompt.

\begin{figure}[t]
    \centering
    \includegraphics[width=0.79\linewidth]{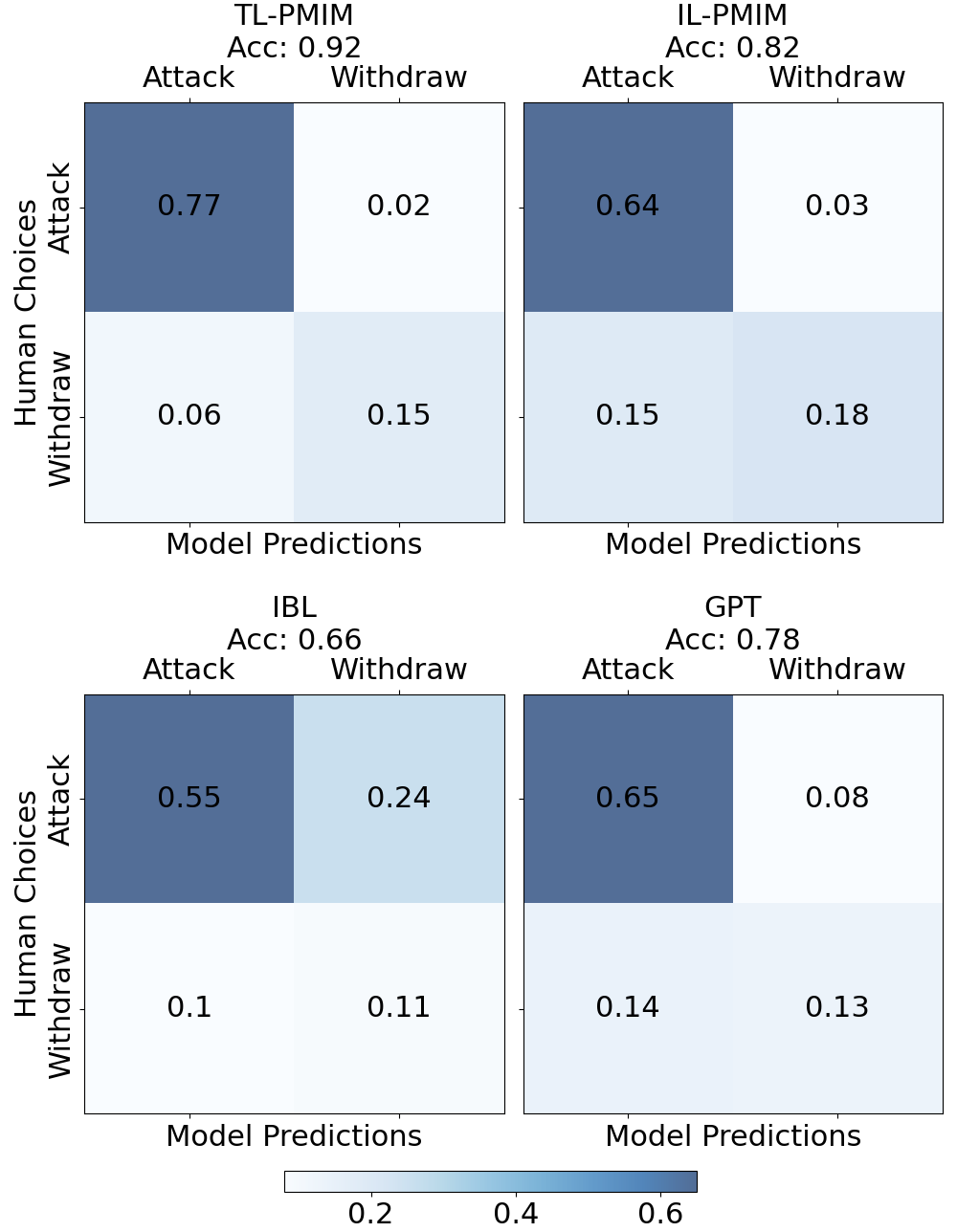}
    \caption{IAG confusion matrix}
    \label{fig:iag}
\end{figure}

\begin{figure*}[t]
    \centering
    \includegraphics[width=0.85\textwidth]{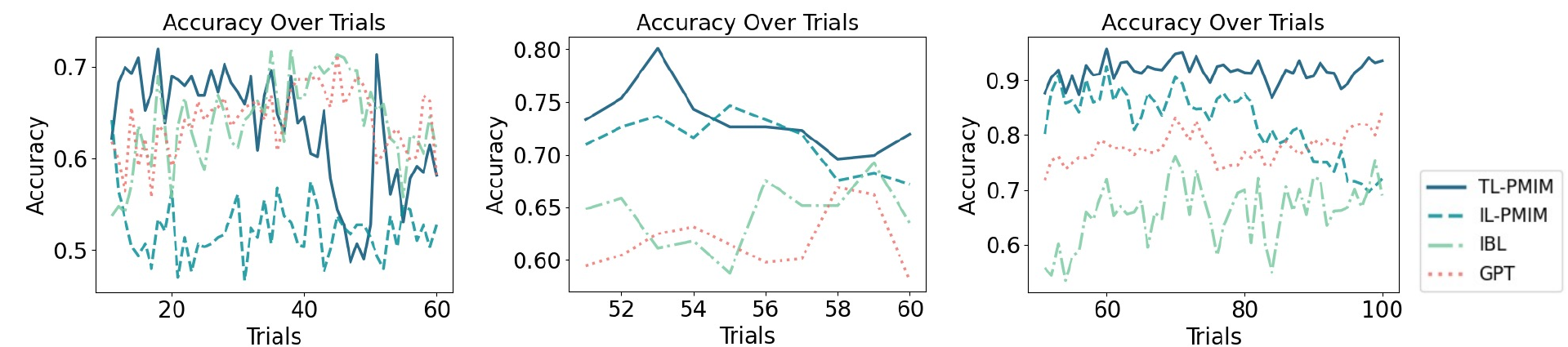}
    \caption{Model-human alignment over trials in the test set. (Left) Phishing 10:50; (Middle) Phishing 50:10; (Right) IAG}
    \label{fig:overtrials}
\end{figure*}


\subsection{Evaluation and Comparison}

\textbf{Neural Networks Increase Fidelity}:
We show the average test performance (accuracy in matching human decision) of the models under consideration in Figures~\ref{fig:phish5010},~\ref{fig:phish1050}, and~\ref{fig:iag}. It can be seen that the performance of TL-PMIM and IL-PMIM drops in the phishing domains when the amount of training data is lower. Supporting this trend, their performances are much better in the high data IAG domain. IBL seems to have similar performance in all domains, possibly due to limited model capacity. GPT does not suffer much in low data domain, possibly due to a much richer prior knowledge base inbuilt into the GPT3.5 model.

\begin{figure*}[t]
    \centering
    \includegraphics[width=0.82\textwidth]{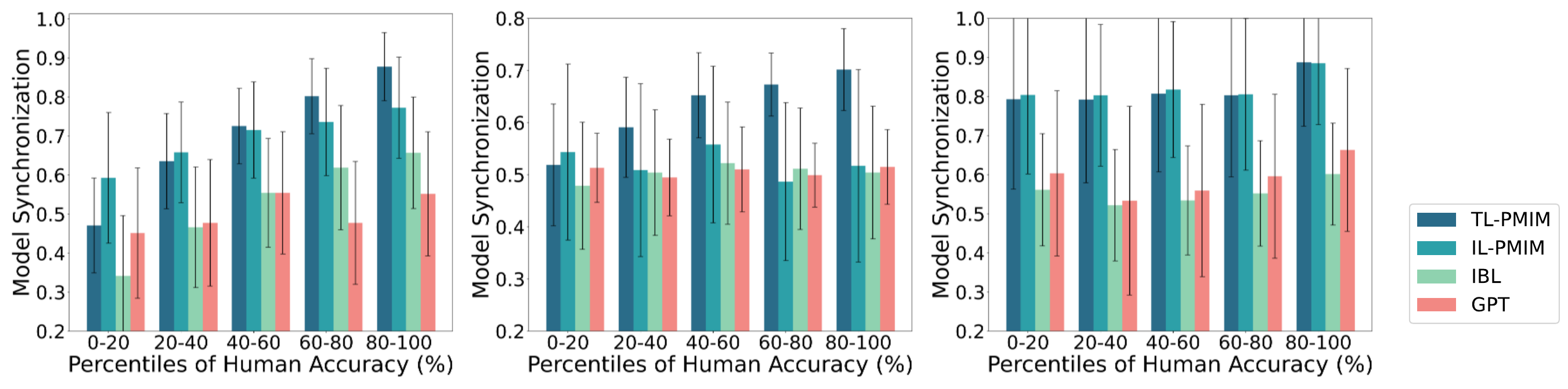}
    \caption{Model performances split by human's accuracy (Left) Phishing 50:10, (Center) Phishing 10:50, (Right) IAG }
    \label{fig:splitground}
\end{figure*}

In more details, Figure~\ref{fig:overtrials} shows the average performance (over participants) of the models on each trial in the test set. For phishing emails, in the 10:50 split we see a drop in performance of all models, except IBL, starting at the 51st trial. This is due to the change in the distribution of emails for the last 10 post-training emails, as stated in the data description. IBL does not suffer a drop in this case, likely because its low capacity leads to better out of distribution performance. 

\textbf{Interpretability}:
\begin{figure}[t]
    \centering
\includegraphics[width=0.9\linewidth]{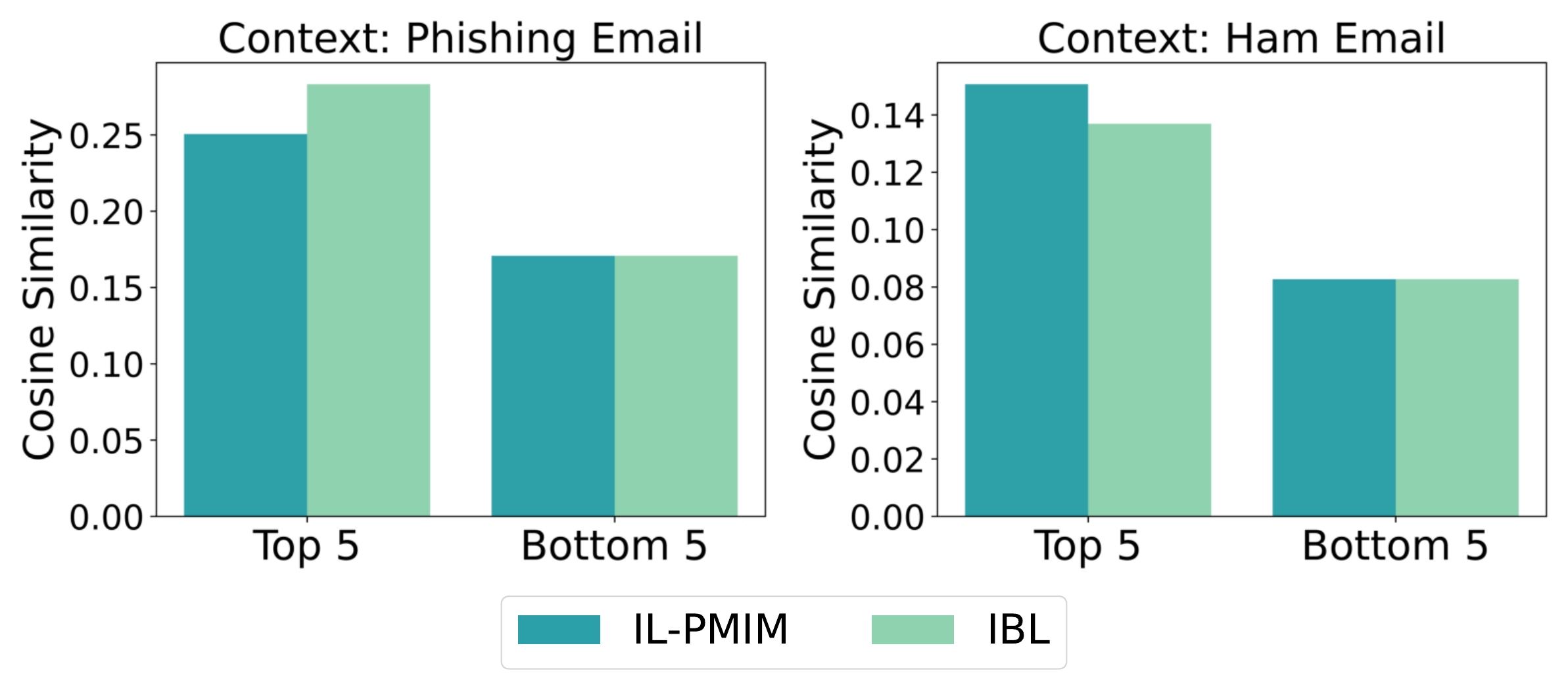}
    \caption{Average similarity between the six auxiliary features of the current email to the top ranked emails in the memory and bottom ranked emails in the memory.}
    \label{fig:cosine}
\end{figure}
Next, we look at which emails in memory do the interpretable models IBL and IL-PMIM pay attention to when deciding about the next email $e_0$. In particular, we can extract the relevance of emails in memory for IBL using the \emph{retrieval probability} defined before Equation~\ref{eq:prob_retrieval}. For IL-PMIM we can do the same using the \emph{attention weights} defined in Equation~\ref{eq:attention}. Then, we rank the emails in memory and extract the top five relevant ones $R$ and the bottom five relevant ones $N$. Recall from the phishing data description, that for every email $e$ we have an auxiliary vector $v_e$ of size six, denoting the presence or absence of characteristics that make the email a phishing email or not. Then, we compute $(1/5)\sum_{e \in R} Sim(v_{e_0}, v_e)$ and $(1/5)\sum_{e \in N} Sim(v_{e_0}, v_e)$ where $Sim$ is cosine similarity. These numbers measure how relevant the top five and bottom five emails are to the current email from the perspective of phishing characteristics.  
We expect to see a difference between these two numbers, which is what we show in Figure~\ref{fig:cosine}. The figure shows the results split by ground truth phishing or ham email and shows that IL-PMIM and IBL have roughly the same degree of attention difference between the top and bottom five cases. {\color{black}An example for both IL-PMIM and IBL is provided in the appendix.} 

Next, we note that TL-PMIM is not interpretable by design, and while there is a lot of work on explaining transformer outputs, the explainability tools generally have no guarantees. We also found that these tools do not scale to the size of our Llama 7B model with some of the longer context lengths. Nonetheless, we provide some qualitative results next for TL-PMIM. using an interpretability tool~\cite{achtibat2024attnlrp} for a phishing email (with small memory length of 10 instances) in the box below; this email was correctly classified by the human and the TL-PMIM matched human decision. The parts highlighted in yellow by the tool are considered important by the model, which can be considered as a proxy for what the human focuses on. In the box, we also mark (as superscript) parts of email that indicate the phishing nature of the email based on the six auxiliary labels we have for all emails (this email has five characteristics). It can be seen that these marked parts are also highlighted.

{\small
\begin{tcolorbox}[colback=white!10, colframe=black, grow to right by=0.1cm,left=0pt,right=0pt,top=0pt,bottom=0pt,title=Interpretability tool~\cite{achtibat2024attnlrp} output]
Sender: accounts@\hl{paypal.security.com}$^5$ \\
Subject: \hl{Un}autho\hl{rized Account Access Attempt}$^1$ \\
Body: \hl{Potential} Unauthorized Account Access \hl{Attempt!} \hl{Case id} : 9000321-128. \hl{Login attempt} from \hl{unknown} device. \hl{Date:} 1/\hl{25/2017} \hl{Time: 12}:00 \hl{PM} EST \hl{Dear Client On 1/25/2017} an attempt was made to access your account from an unknown \hl{IP Address} (54.875.24.152). In an \hl{effort} to protect your personal and financial \hl{information we have temporarily}$^2$ locked your account. This \hl{includes} the ability to withdraw funds. ...[Abridged to fit in main paper]...
You \hl{must} complete the security check \hl{within 72 hours}$^2$ of the \hl{date} of this \hl{email}. Should you choose not to \hl{comply, your} account \hl{will} be \hl{permanently locked} and any funds contained therein will be forfeited. To \hl{begin, please: 1. Right click on}$^3$ the \hl{link PayPal Member Security Services}$^3$ 2. \hl{Open} the \hl{page} in a \hl{secure (private) browser window}. 3. Follow the instructions provided. \hl{Please} note \hl{that PayPal} will never ask for your \hl{password}$^4$ over the phone, nor ask you to send your login credentials via email. \hl{Don't wait}$^2$. \hl{Protect} your information and unlock your account within the \hl{next 72 hours to}$^2$ avoid the \hl{loss} of \hl{your} account and \hl{forfeiture of funds}. \\

Expert Labels: ($1$) Subject suspicious, ($2$) Asks to do something urgently, ($3$) Mismatch in link text and location, ($4$) Requests Credentials, ($5$) Sender Mismatch 
\end{tcolorbox}
}


\textbf{More Fine-grained Analysis}:
We note that \textit{prediction accuracy increases with human's accuracy}.
We split the results by how accurate the human prediction of the ground truth is, which in the phishing domain is whether the email was actually phishing or not and in IAG whether it was optimal to continue the attack or not. We split the result of model-human decision synchronization by the accuracy of the ground truth prediction by humans and show the results in Figure~\ref{fig:splitground}. The number of participants in each bucket in the figure is roughly the same. We notice that with sufficient data (phishing 50:10 or IAG), the models generally do better when the human decisions align with the ground truth. This is not true for GPT which has similar performance in most scenarios, suggesting it relies on its prior knowledge more than trying to mimic the human's observed performance. 

Further, for the phishing 10:50 split, TL-PMIM shows improvement similar to the sufficient data scenario; this further suggests that fine-tuning an LLM helps in better mimicking human behavior, especially when the human behavior exhibits learnable patterns such as being able to detect phishing email. We also conjecture that when the human does not predict the ground truth well, then there may be a mix of various suboptimal behaviors by a single human. This is likely why for the low data domain of phishing (per participant) the performance at lower human accuracy is much worse than in the high data domain of IAG. 

\begin{figure}[t]
    \centering
    \includegraphics[width=0.95\linewidth]{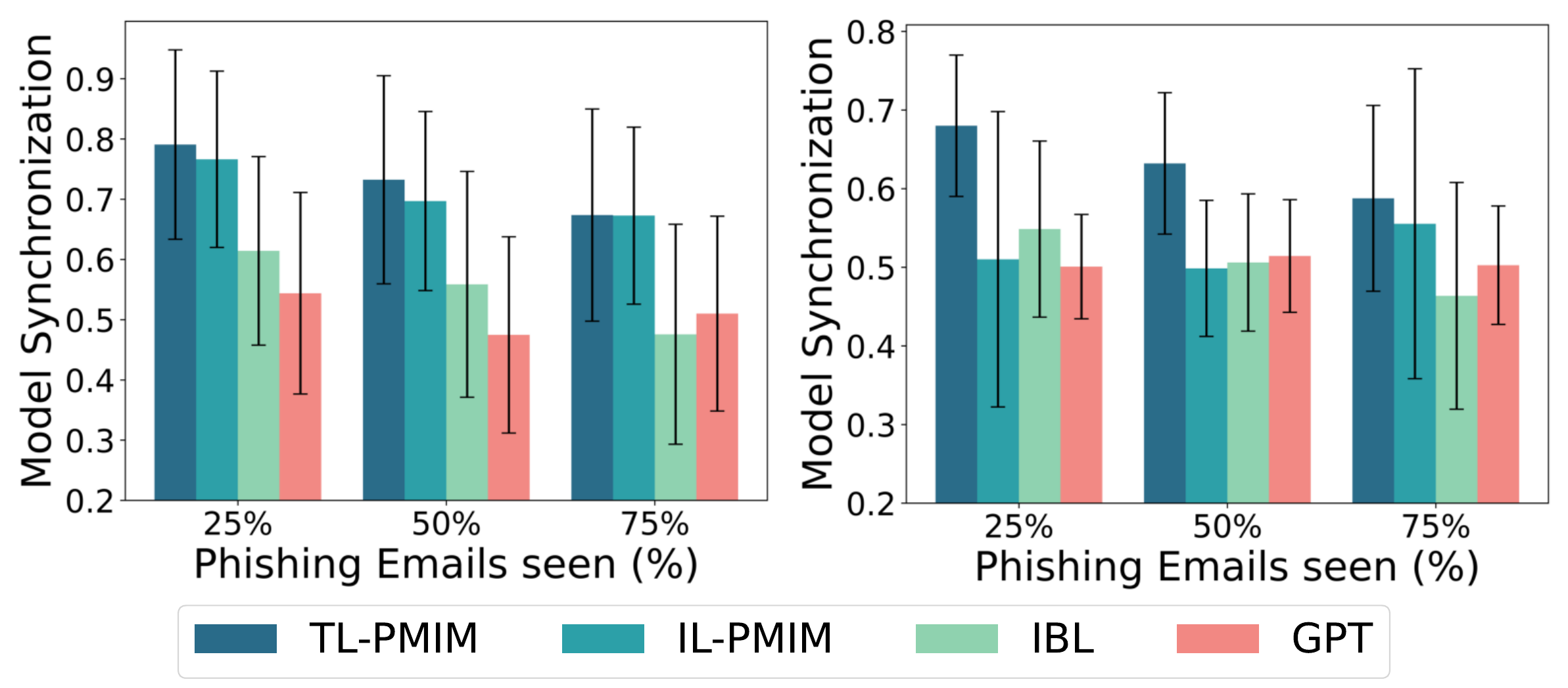}
    \caption{Model performances split by \% of phishing (Left) Phishing 50:10, (Right) Phishing 10:50}
    \label{fig:percentagephish}
\end{figure}

Next,
Figure~\ref{fig:percentagephish} shows the results in the phishing domain split by participants who see different percentages of phishing emails in training. All models, except GPT, do slightly better in 50:10 split for those participants that see fewer phishing emails, given that we observed that overall human accuracy (of detecting phishing and ham emails) is slightly higher when they see fewer phishing emails. The variance across the different percentage groups is minimal for the 10:50 split. 
We present yet more fine-grained results in the appendix where we look at the results split by different labels of ground truth.

\section{Conclusion and Limitations}\label{sec:limit}

Our work aims to show how neural networks can enrich cognitive modeling of dynamic decision-making.
We note a limitation that is generally true for most human subject studies. We showed our results with data collected in a structured lab setting for real world tasks where the environment can be very complex. E.g., when detecting phishing emails in an office setting, humans can be juggling many other tasks, and therefore factors such as the time available to make the decision become important; these have been considered in cognitive science~\citep{kerstholt1995decision}. It is always possible to expand our neural network models with these additional factors (if measured and recorded) but such a work is beyond the scope of a single paper.
 Overall, our work has shown that there is positive synergy to be gained from bringing together concepts from modern deep learning and human cognition sciences; we envision neural networks being adopted extensively in cognitive sciences for higher-fidelity modeling of human behavior in complex tasks.
 
 
\section{Acknowledgments}
This research was sponsored by the Army Research Office and accomplished under MURI Grant Number W911NF-20-S-000.

{
\small
\bibliography{aaai25}
}

\newpage
\onecolumn
\appendix
\renewcommand\thefigure{A.\arabic{figure}}
\renewcommand\thetable{A.\arabic{table}}
\setcounter{figure}{0}
\setcounter{table}{0}

\section{Prompt Templates Used}
\begin{tcolorbox}[colback=gray!10, colframe=black, title=Prompt used in TL-PMIM for the Phishing task]
    Do you think this email is phishing? (A list of history emails and decisions are included):\\\\
    
    YES\\
    NO\\\\
    
    \#\# History Emails and Responses:\\
    \# email 1\\
    Sender: no-reply@yahoo.com\\
    Subject: Password change for your Yahoo account\\
    Body: Hi Ethan, The password for your Yahoo account, a er, was recently changed. If you made this change, you're all set. Didn't change your password? Please take these steps to secure your account. Recover your account here:https://login.yahoo.com/forgot? Review your phone numbers and email addresses and remove the ones that don't belong to you.https://login.yahoo.com/account/security? Thanks, Yahoo\\
    Response: NO\\
    Feedback: Null.\\\\
    
    \# email 2\\
    Sender: message@adobe.com\\
    Subject: Your Adobe password has been reset\\
    Body: Success! Hi Patricia, You have successfully reset the password for your Adobe ID and Creative Cloud account. Continue to your account please click on the following link. Adobe Thank you, The Creative Cloud team\\
    Response: NO\\
    Feedback: Null.\\\\
    
    (ignore the content due to space limit ...)\\\\
    
    \# email 13\\
    Sender: service@remitly.com\\
    Subject: Your Remitly Account has been deleted\\
    Body: Greetings from Remitly.com, As you requested, we have deactivated your Remitly account. We appreciate your past business and we look forward to you coming back soon. If you want to activate your account again please Contact Customer Service to reactivate your account . Thank you for using our services. - The Remitly Team\\
    Response: NO\\
    Feedback: This email is ham.\\\\
    
    \#\# Current Email:\\
    Sender: jim.jenkins@global-infra.com\\
    Subject: Revised Desktop Support Service\\
    Body: The executive desktop support services have been streamlined due to recent organizational changes. Click here to view the revised service offerings. We appreciate your cooperation in this matter. Global Infrastructure Team\\\\

    \#\# Response: \\

\end{tcolorbox}

\begin{tcolorbox}[colback=gray!10, colframe=black, title=Prompt used in TL-PMIM for the IAG task]
   You are playing the role of employees at a company, with the objective to maximize your score by 'hacking' computers to steal proprietary information. Two security analysts are tasked with monitoring the computers. Points are earned by successfully avoiding detection, and points are lost if you are caught. Based on the provided information about the current status, you need to decide the next action: 0 (withdraw attack) or 1 (continue attack), in order to maximize the score. Feedback on your action is provided immediately. The features include:

   \begin{itemize}
       \item [--] TargetNum: An integer from 1 to 6, representing the ID number of the target computer.
       \item [--] Location: An integer from 1 to 6 (1 = top-middle, 2 = top-right, 3 = bottom-right, 4 = bottom-middle, 5 = bottom-left, 6 = top-left), representing the location ID of the selected target.
       \item [--] Payment: The potential reward value if the attack is successful.
       \item [--] Penalty: The potential penalty value if the attack is unsuccessful.
       \item [--] Mprob: The probability that the selected target is being monitored by security analysts.
       \item [--] Warning: A binary flag where 0 indicates no warning signal presented, and 1 indicates a warning signal was presented to the attacker.
       \item [--] Feedback: -10 to 10, The value of the reward/penalty given the attacker's action and target coverage. Penalty if the target covered else Payment, or 0 if withdrew attack.
   \end{itemize}

    \#\# History attacks:\\
    \# attack 1
    \begin{itemize}
        \item [--] Features (TargetNum, Location, Payment, Penalty, Mprob, Warning): (1, 2, 2, 1, 0.22, 0)
        \item [--] Action: 1
        \item [--] Feedback: 2
    \end{itemize}

    \# attack 2
    \begin{itemize}
        \item [--] Features (TargetNum, Location, Payment, Penalty, Mprob, Warning): (6, 1, 5, 5, 0.36, 0)
        \item [--] Action: 1
        \item [--] Feedback: 5
    \end{itemize}
    
    \#\# Current attack:
    \begin{itemize}
        \item [--] Features (TargetNum, Location, Payment, Penalty, Mprob, Warning): (6, 1, 5, 5, 0.36, 0)
    \end{itemize}
    
    \#\# Action: \\
    
\end{tcolorbox}

\begin{tcolorbox}[colback=gray!10, colframe=black, title=Prompt used in GPT3.5 for the phishing task]

Your objective is to anticipate the human's classification of the current email by leveraging their historical responses to similar emails. The cognitive model guiding this prediction asserts that the individual's reactions are exclusively shaped by their prior interactions with comparable email content. Additionally, ensure your decisions are substantiated by referencing the relevant historical emails and responses to provide a comprehensive justification for your predictions. \\

\#\# Example \\
Consider the following example: \\

\#\#\# History Emails and Human Responses: \\
Email 1: \\
Subject: Unusual Login Activity Detected \\
Body: [Content] \\
Response: YES \\
Sender: support@yourbankonline.com \\

Email 2: \\
Subject: Account Security Update Required \\
Body: [Content] \\
Response: YES \\
Sender: noreply@shoppingwebsite.com \\

Email 3: \\
Subject: Exclusive Limited-Time Offer \\
Body: [Content] \\
Response: NO \\
Sender: security@onlineinvestment.com \\

\#\#\# Current Email and Human Response: \\

Subject: Immediate Attention Required - Possible Unauthorized Transaction \\
Body: [Content] \\
Justification: Similar to Email 1 from the history, where the sender was 'support@yourbankonline.com' with a subject about 'Account Security Update Required,' the current email exhibits comparable characteristics. The historical response pattern suggests that emails from addresses resembling official bank sources, coupled with urgent security concerns, were consistently labeled as phishing (YES) by the human. \\
Response: YES \\

\#\# Email of interest: \\

Apply this rationale to predict the response for the current email. \\
$<$Current Email$>$
\end{tcolorbox}

\begin{tcolorbox}[colback=gray!10, colframe=black, title=Sample response of GPT3.5 in the phishing task]
Justification: \\
Based on the historical responses provided, the individual consistently marked emails related to offers and promotions as 'NO.' In Email 3, the subject line was 'Exclusive Limited-Time Offer' from the sender 'security@onlineinvestment.com,' and the response was 'NO.' Similarly, the current email from 'discount@eddiebauer.com' with the subject line 'STARTS TODAY! 40\% Off Everything!' falls under the category of promotional content, which has been consistently classified as 'NO' in past interactions. \\
Response: NO
\end{tcolorbox}

\begin{tcolorbox}[colback=gray!10, colframe=black, title=Prompt for GPT3.5 in the IAG task, breakable]
Your goal is to predict the actions of a human in a certain task. The cognitive model guiding this prediction asserts that the individual's reactions are exclusively shaped by their previous actions in similar instances.

\#\#   Task Description \\
You are an employee participating in a high-stakes corporate espionage operation, tasked with infiltrating computers to steal critical proprietary information while evading detection by security analysts. Each hacking attempt presents a calculated risk, with rewards for successful breaches and consequences for detection. Each attempt occurs in two stages, 1) target selection: decide which computer to attack, after this, you may receive a signal from the system stating whether it is being monitored. Note that this signal maybe a bluff, i.e, it may claim it is being monitored while it is not, and 2) attack/withdraw: decide whether to continue attacking the target. Feedback on the status of the attack is provided immediately. \\
Below are the features of each instance and the action space of the human. \\

\# Features \\
- TargetNum: An integer from 1 to 6, representing the ID number of the target computer. \\
- Location: An integer from 1 to 6 (1 = top-middle, 2 = top-right, 3 = bottom-right, 4 = bottom-middle, 5 = bottom-left, 6 = top-left), representing the location ID of the selected target. \\
- Payment: The reward if the attack is successful. \\
- Penalty: The penalty if the attack is unsuccessful. \\
- Mprob: The probability that the selected target is being monitored by security analysts. \\
- Warning: A binary flag where 0 indicates no warning signal presented, and 1 indicates a warning signal was presented to the attacker. \\
- Feedback: -10 to 10, The value of the reward/penalty given the attacker's action and target coverage. Penalty if the target covered else Payment, or 0 if withdrew attack. \\

\# Actions \\
- Withdraw: 0 \\
- Attack: 1 \\

Based on past actions of the human in similar instance, your task is to predict the SECOND decision only (whether to attack/withdraw). Additionally, ensure your decisions are substantiated by referencing the relevant instances in the history to provide a comprehensive justification for your predictions. \\

\#\# Example \\
Consider the following example: \\

\#\# History attacks: \\
\# attack 1\\
- Features (TargetNum, Location, Payment, Penalty, Mprob, Warning): (1, 2, 2, 1, 0.22, 0)\\
- Action: 1\\
- Feedback: 2\\
\# attack 2\\
- Features (TargetNum, Location, Payment, Penalty, Mprob, Warning): (6, 1, 5, 5, 0.36, 0)\\
- Action: 1\\
- Feedback: 5\\
\# attack 3\\
- Features (TargetNum, Location, Payment, Penalty, Mprob, Warning): (4, 3, 9, 10, 0.4, 0)\\
- Action: 1\\
- Feedback: 9\\

\#\# Current attack:\\
- Features (TargetNum, Location, Payment, Penalty, Mprob, Warning): (4, 1, 8, 9, 0.37, 1)\\

\#\# Justification: \\
The human chose to attack the same location in the past and target with similar Mprob, both leading to succesful outcomes. Hence, it is likely that the human will continue with this approach. \\

\#\# Action: 1 \\

\#\# Current instance \\
Predict the decision on the following instance: \\
$<$Current Instance$>$

\end{tcolorbox}

\begin{tcolorbox}[colback=gray!10, colframe=black, title=Sample response of GPT3.5 in the IAG task]
Justification: \\
Based on the historical data, the human consistently attacked the same target location (Location 4) in the past with mostly successful outcomes. Additionally, the current target has a lower probability of being monitored compared to previous instances. Therefore, it is likely that the individual will choose to attack the current target. \\
Action: 1
\end{tcolorbox}

\newpage 
\section{More Results}
\begin{figure}[h]
    \centering
    \includegraphics[width=0.7\textwidth]{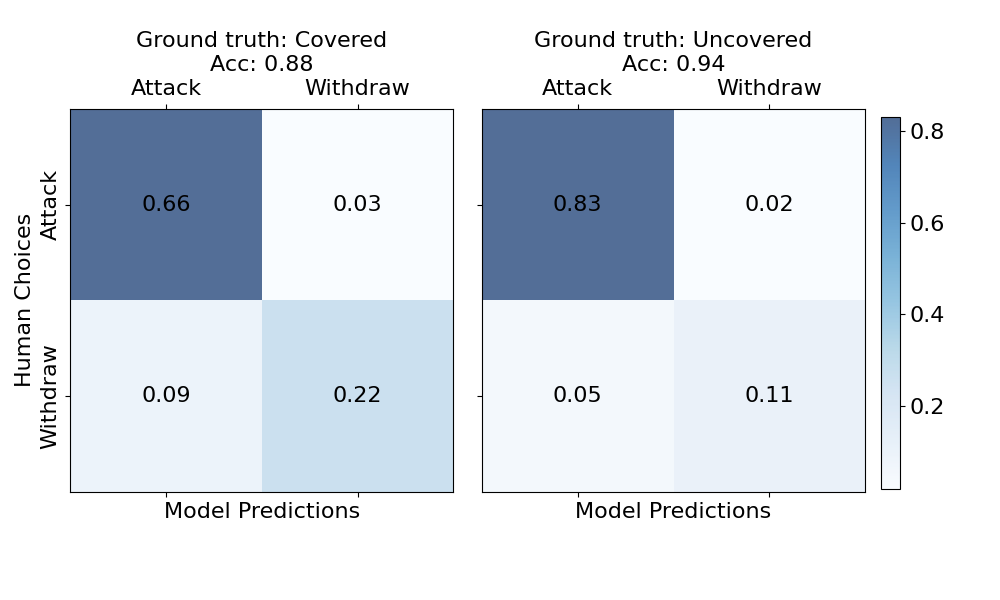}
    \caption{IAG/TL-PMIM}
    \label{fig:appendixfirst}
\end{figure}

\begin{figure}[h]
    \centering
    \includegraphics[width=0.7\textwidth]{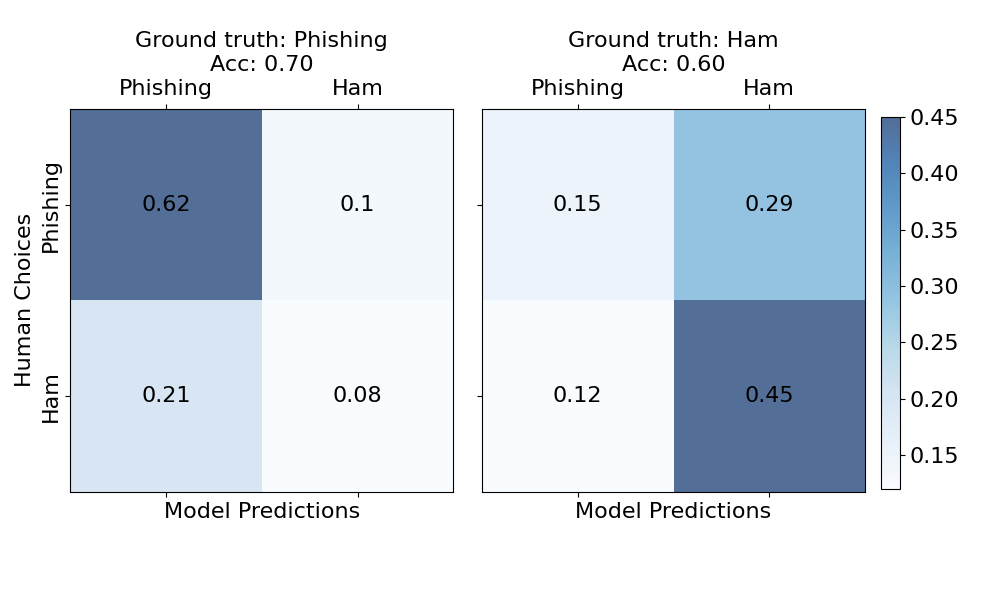}
    \caption{phishing-1050/GPT}
    \label{fig:appendixlast}
\end{figure}

Here we provide more aggregate results split by the ground truth for all the domains. The ground truth for IAG is labeled as covered or uncovered, which directly corresponds to whether the optimal decision should be to not continue with attack or continue with attack. These are shown in 
Figures~\ref{fig:appendixfirst}-\ref{fig:appendixlast}.

\begin{figure}[h]
    \centering
    \includegraphics[width=0.7\textwidth]{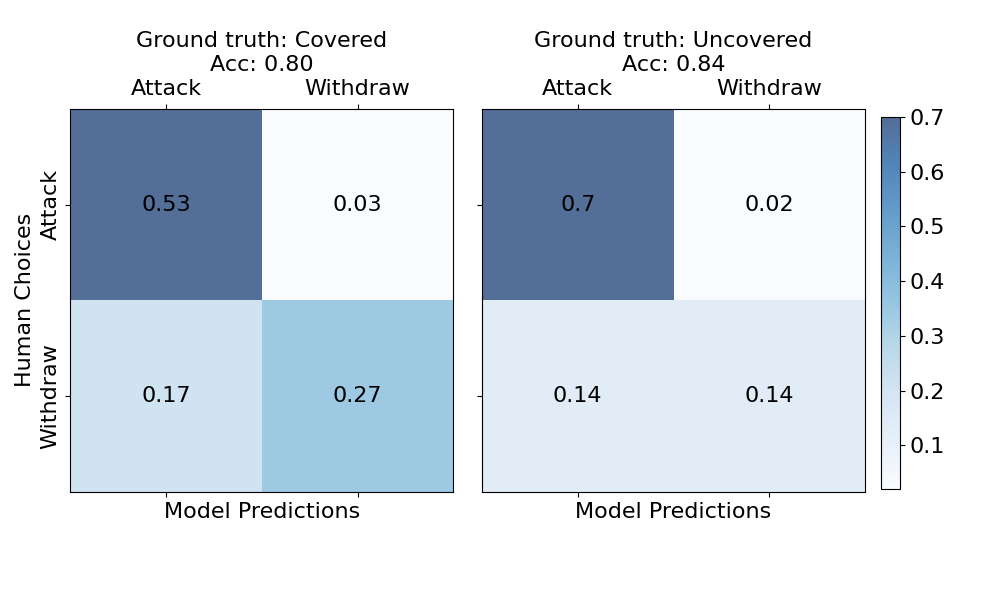}
    \caption{IAG/IL-PMIM}
    \label{}
\end{figure}

\begin{figure}[h]
    \centering
    \includegraphics[width=0.7\textwidth]{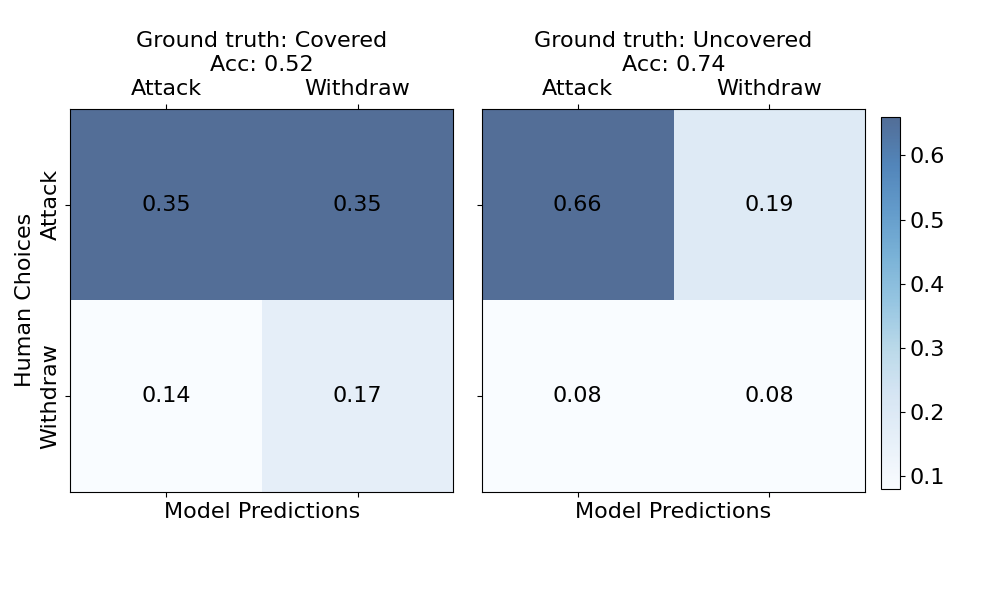}
    \caption{IAB/IBL}
    \label{}
\end{figure}

\begin{figure}[h]
    \centering
    \includegraphics[width=0.7\textwidth]{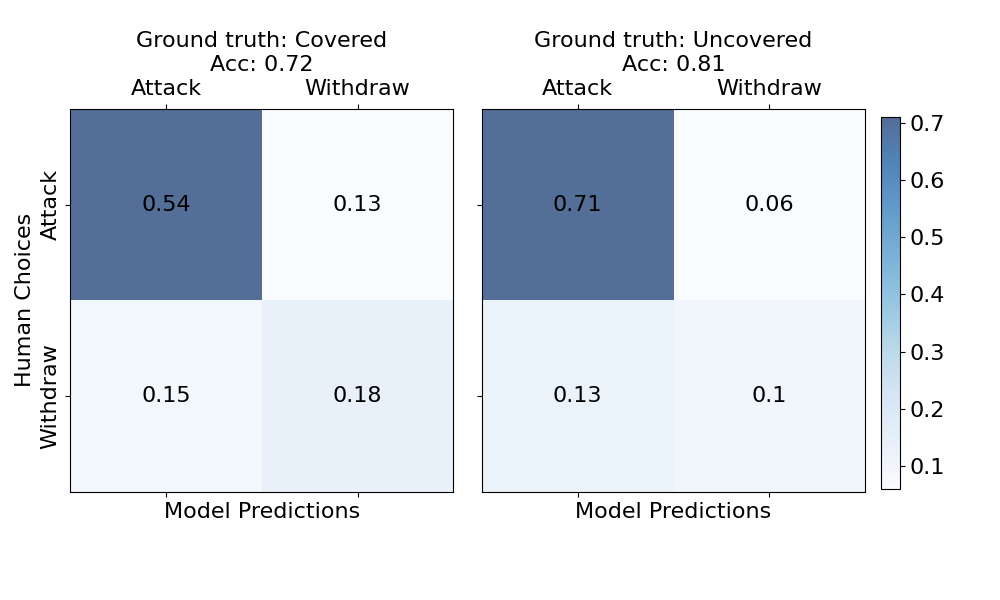}
    \caption{IAG/GPT}
    \label{}
\end{figure}

\begin{figure}[h]
    \centering
    \includegraphics[width=0.7\textwidth]{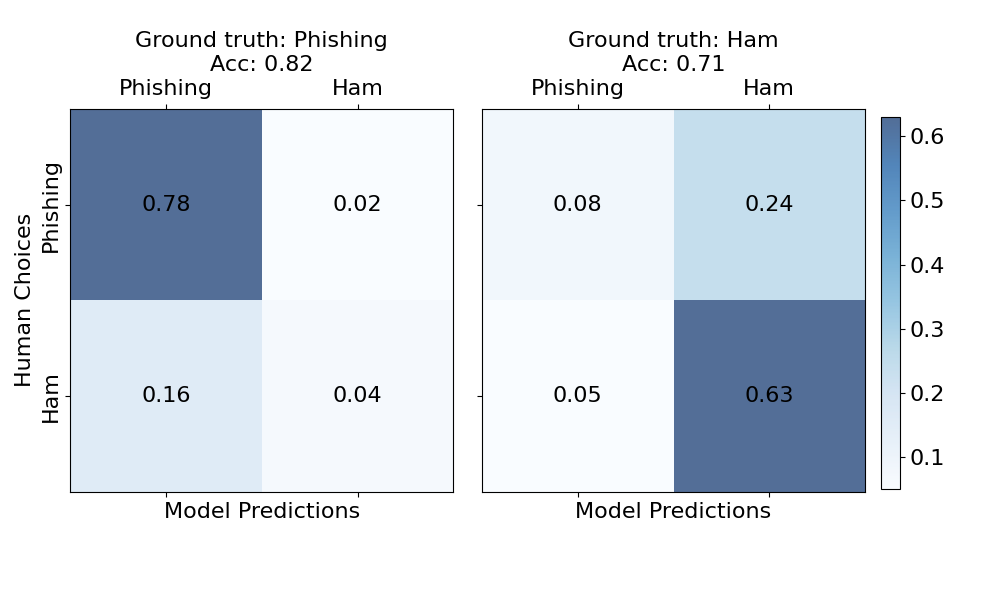}
    \caption{phishing-5010/TL-PMIM}
    \label{}
\end{figure}

\begin{figure}[h]
    \centering
    \includegraphics[width=0.7\textwidth]{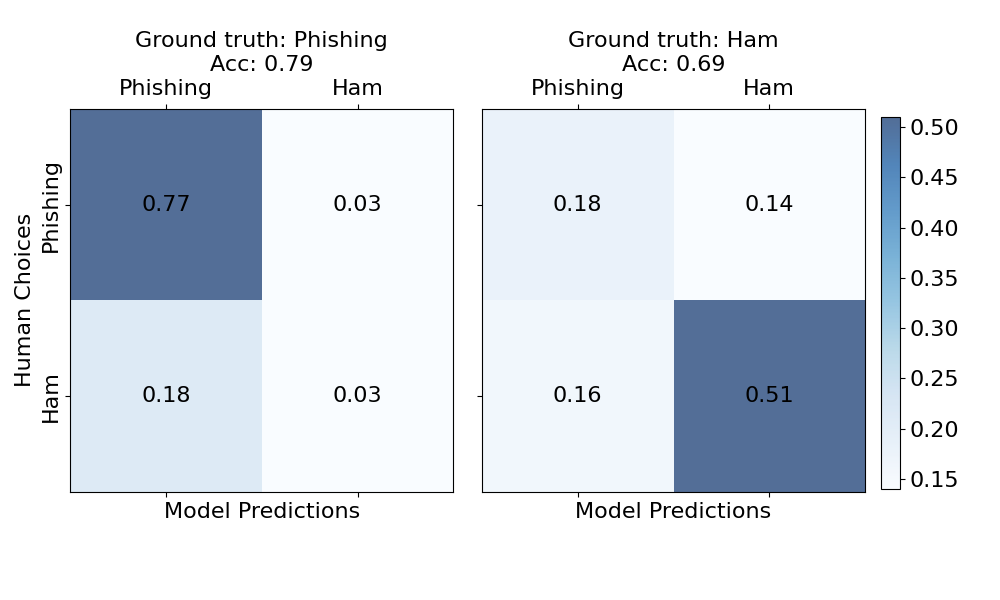}
    \caption{phishing-5010/IL-PMIM}
    \label{}
\end{figure}

\begin{figure}[h]
    \centering
    \includegraphics[width=0.7\textwidth]{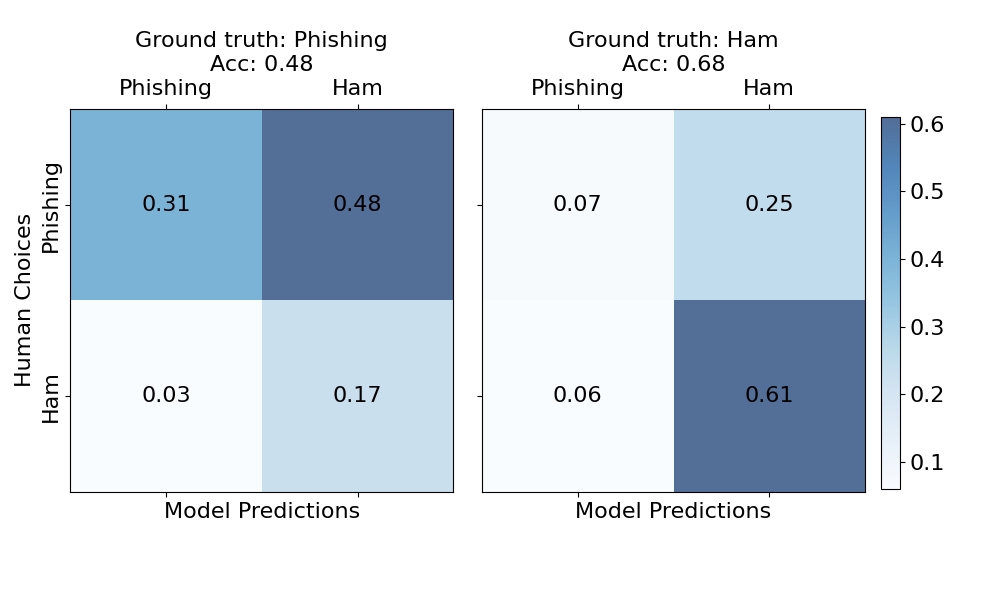}
    \caption{phishing-5010/IBL}
    \label{}
\end{figure}

\begin{figure}[h]
    \centering
    \includegraphics[width=0.7\textwidth]{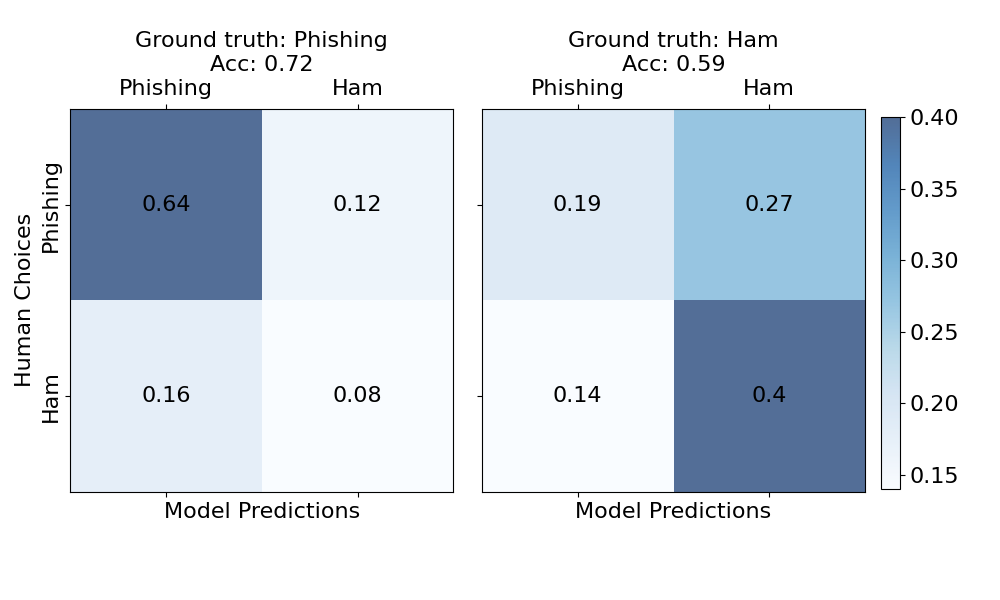}
    \caption{phishing-5010/GPT}
    \label{}
\end{figure}

\begin{figure}[h]
    \centering
    \includegraphics[width=0.7\textwidth]{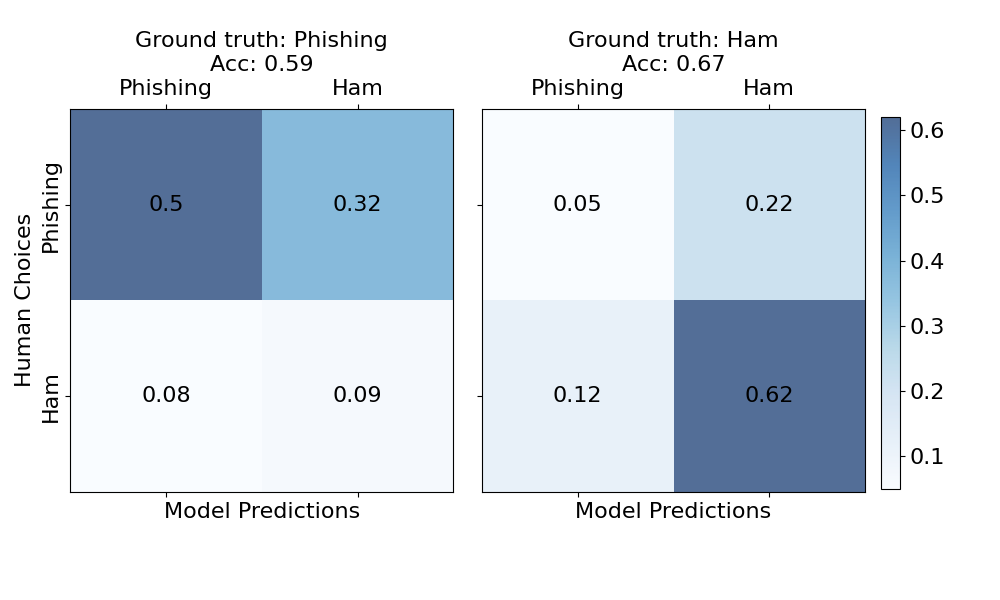}
    \caption{phishing-1050/TL-PMIM}
    \label{}
\end{figure}

\begin{figure}[h]
    \centering
    \includegraphics[width=0.7\textwidth]{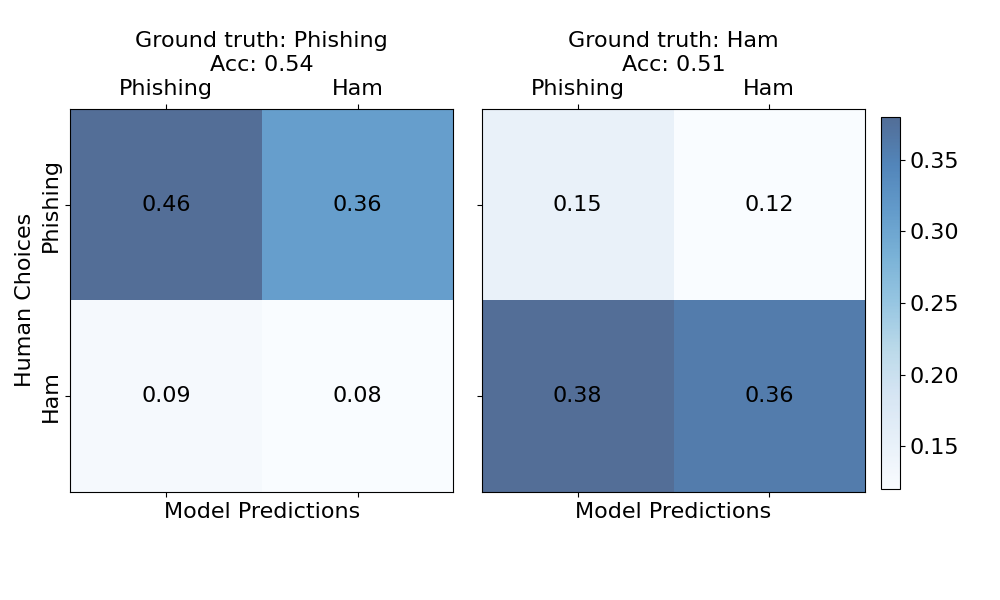}
    \caption{phishing-1050/IL-PMIM}
    \label{}
\end{figure}

\begin{figure}[h]
    \centering
    \includegraphics[width=0.7\textwidth]{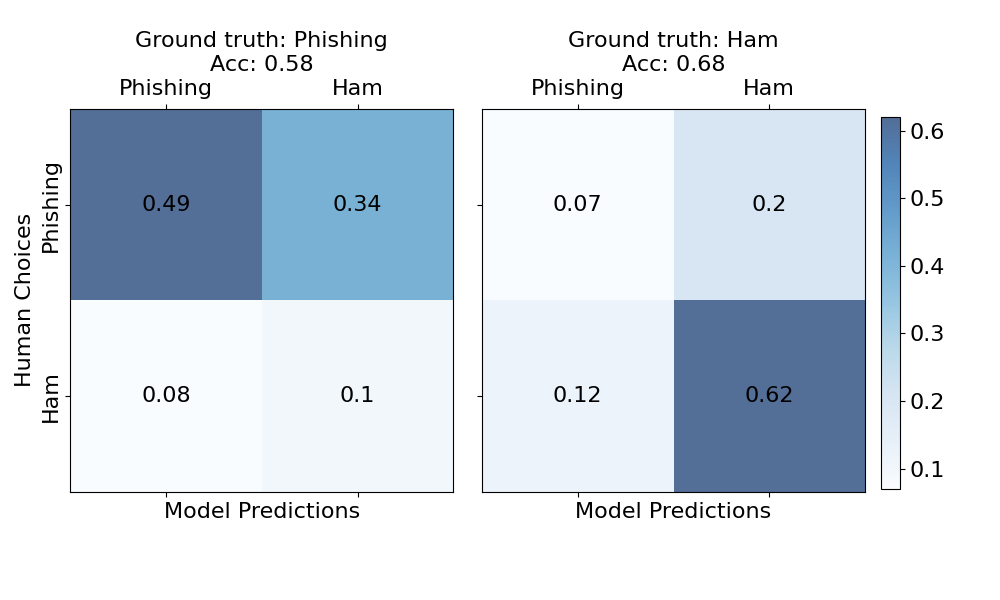}
    \caption{phishing-1050/IBL}
    \label{}
\end{figure}

\clearpage
\newpage

\section{Additional IBL Details}
While IBL is a baseline for our work, we performed some tuning. We conducted model fitting to determine the optimal decay parameter for each participant, taking into account previous research indicating that effective adjustment to choice environments is linked to higher decay levels ($d$)~\cite{konstantinidis_patterns_2022}. This suggests that a weaker memory may actually lead to better adaptation. Our decay search involved testing decay values ranging from $0.1$ to $3$, in increments of $0.01$.
We analyzed the synchronization of model predictions with participant decisions for each of the models generated (291 models) for each participant. At each trial $t$, we assigned a value of $1$ when the model prediction aligned with the human action and a value of $0$ when it did not for each participant. Subsequently, we calculated the proportion of synchronization, which we call SyncRate, for each model and participant. The model exhibiting the highest SyncRate was selected as the best fit for the participant's data. In cases where multiple SyncRates had the same highest value, the highest decay value was chosen.

The IBL model is computationally cheap and ran on a CPU of 1.8GHz speed and 16GB RAM.

{\color{black}\section{Interpretability of IL-PMIM and IBL as Retrieved Examples}

We provide the instances with the \emph{highest contribution} to the final decision for both the IBL model and IL-PMIM for the current email shown below and the memory content exactly the same for both models. Although these top-contributing instances may not share semantic similarity with the current instance, they exhibit similarities in expert labels, indicating that they possess similar phishing characteristics.}

\begin{tcolorbox}[colback=gray!10, colframe=black, title=Current Instance (Phishing Task)]
Sender: david@gmail.com\\
Subject: Business Receipt\\
Body: To Whom It May Concern, I recently did business with your company, and I had a question about my receipt. I'm just confused about why I was charged this extra \$250 from your guys? I don't remeber that being in the cost. I would like this sorted as ASAP! My Receipt https://drive2.google.com/file/ dj9efVJ4jj0g Thank you! Your Loyal Customer, David Grey
\\
\\
Ground Truth:  phishing\\
Features:  [0 0 0 1 0 1]
\\
\\
*The \textbf{Features} represent human expert annotations, where 1 indicates the presence of a specific phishing characteristic, and 0 indicates its absence. The phishing features include: [sender\_mismatch, request\_credentials, subject\_suspicious, urgent\_ask, offer, link\_mismatch]
\end{tcolorbox}

\begin{tcolorbox}[colback=gray!10, colframe=black, title=Retrieved Example in IBL Model]
Sender: julia@dinkle.einstein.com\\
Subject: Possible Inheritance\\
Body: Hello! This letter is to inform you of a possible inheritance left in your name by someone you may have known who is now deceased. You are listed as a close relative to the departed. We need to verify important details with you before we continue any further. This matter is time sensitive. We must receive a reply within 24 hours or your inheritance may be forfeited. Further Details Here Thank you. Sincerely, Julia Carraba Dinkle \& Einstein, LLC
\\
\\
Human Response:  phishing\\
Ground Truth:  phishing\\
Features:  [1 1 1 1 1 1]
\end{tcolorbox}

\begin{tcolorbox}[colback=gray!10, colframe=black, title=Retrieved Example in IL-PMIM]
Sender: support@security.amazon.com\\
Subject: Your Amazon Account \\
Body: The account number associated with your email address has been suspended. Account Number : 108-4596473-8009841 FF Dear Valued Customer, We were unable to validate important details about your Amazon Shopping account. Your Amazon Shopping account has been suspended until all verification steps have been completed. We are sorry for any inconvience and we look forward to having you as a customer again, thank you for you prompt attention to this matter. Please visit your account details to confirm the payment information for your account. Verify Account Information Account Details Account \#108-4596473-8009841 Amazon IT Team Conditions of Use Privacy Notice 1996-2017, Amazon.com, Inc. or its affiliates 05/10/2017
\\
\\
Human Response:  phishing\\
Ground Truth:  phishing\\
Features:  [1 1 0 1 0 1]
\end{tcolorbox}

\end{document}